\documentclass[sigconf]{acmart}

\AtBeginDocument{%
  \providecommand\BibTeX{{%
    \normalfont B\kern-0.5em{\scshape i\kern-0.25em b}\kern-0.8em\TeX}}}

\acmConference[KDD 2020]{2020 KDD Workshop on Applied Data Science for Healthcare}{Aug 24, 2020}{San Diego, CA}

\begin{document}

\title{Clinical Recommender System: Predicting Medical Specialty Diagnostic Choices with Neural Network Ensembles}

\author{Morteza Noshad$^{1}$, Ivana Jankovic$^{2}$, Jonathan H. Chen$^{1,3}$}
\email{<noshad>, <ijankovic>, <jonc101>@stanford.edu}


\affiliation{%
  \institution{$^1$Stanford Center for Biomedical Informatics Research, Stanford University, Stanford, CA\\ 
  $^2$Division of Endocrinology, Stanford University School of Medicine, Stanford, CA\\
$^3$Division of Hospital Medicine, Stanford University, Stanford, CA}
}


\begin{abstract}
  
  The growing demand for key healthcare resources such as clinical expertise and facilities has motivated the emergence of artificial intelligence (AI) based decision support systems. 
  We address the problem of predicting clinical workups for specialty referrals. As an alternative for manually-created clinical checklists, we propose a data-driven model that recommends the necessary set of diagnostic procedures based on the patients' most recent clinical record extracted from the Electronic Health Record (EHR). This has the potential to enable health systems expand timely access to initial medical specialty diagnostic workups for patients. The proposed approach is based on an ensemble of feed-forward neural networks and achieves significantly higher accuracy compared to the conventional clinical checklists.
  
  
  

\end{abstract}



\maketitle

\section{Introduction}

The growing limitations in the scarcest healthcare resource - clinical expertise - is
an issue that has long been at the front line of the healthcare industry.
This shortage of clinician experteise is particularly acute in access to medical specialty care. In some locations, patients wait several months for outpatient specialty consultation visits, which contributes to the $20\%$ higher mortality in the US \cite{ prentice2007delayed}.
However, potential solutions have been slow to come.

Our goal is to develop a radically different paradigm for specialty consultations by developing a tier of automated guides that proactively enable initial workup that would otherwise be delayed awaiting an in-person visit. We focus on recommending the clinical orders for medications and diagnostic tests from outpatient consultations that any clinician could initiate with adequate support. This system can consolidate specialty consultation needs and open greater access to effective care for more patients.
A key scientific barrier to realizing this vision is the lack of clinically acceptable tools powered by robust methods for collating clinical knowledge, with continuous improvement through clinical experience, crowdsourcing, and machine learning. Existing tools include electronic consults that allow clinicians to email specialists for advice, but their scale remains constrained by the availability of human clinical experts. Electronic order checklists (order sets) are in turn limited by the effort to maintain and adapt content to individual patient contexts \cite{middleton2016clinical}.


Machine learning approaches are revolutionizing various healthcare areas such as medical imaging\cite{giger2018machine}, diagnostic models \cite{choi2016doctor, miotto2016deep} and virtual health assistants\cite{kenny2008virtual} by introducing more accurate, low cost, fast and scalable solutions.
Automated diagnostic workflow recommendation is another emerging application of machine learning which has so far mainly been focused on predicting the need for specific medical imaging \cite{merdan2018integrating}. However, only a few previous studies have explored the possibility of using machine learning approaches to design a scalable intelligent system that can recommend diagnostic procedures of any type to the patients, as an alternative to the conventional clinical checklists. 
Authors in \cite{lakshmanaprabu2019online} and \cite{chen2017predicting} apply recommender systems based on probabilistic topic modeling and neural networks to predict inpatient clinical order patterns. Other than predicting workflows, recommender systems have also been used for diagnosis in several previous papers \cite{komkhao2013recommender, hao2016comparative}.


In this work, we address the problem of predicting outpatient specialty workflows. Specifically, our objective is to predict which procedures would be ordered at the first specialty visit for a patient referred by a primary care physician (PCP), based on their medical records.
This procedure could provide automated decision support and recommendations at primary care visits or specialist pre-visit screenings to allow diagnostic procedures to be completed while the patient is awaiting their in-person specialist visit.
As opposed to manually-created medical checklists, which are mainly based on diagnosis (e.g., common laboratory and imaging tests a clinician can order to evaluate diabetes), the proposed data-driven algorithm utilizes the patient's previous lab results, diagnosis codes, and the most recent procedures as input and recommends follow-up lab orders and procedures. 
The proposed recommender model offers several key features such scalability, to answer unlimited queries on-demand; maintainability, through automated statistical learning; adaptability to respond to evolving clinical practices; and personalizability of individual suggestions with greater accuracy than manually-authored checklists. 
We categorized the input EHR data into three groups:  diagnostic data, including the diagnosis codes and lab results; procedures ordered by the referring PCP; and the specialist being referred to (recognized by their ID). This grouping of the data lets us use appropriate base models for each of the input data categories and process them separately. 
The first base model is a neural network based multi-label classifier with diagnostic data as input and specialty procedures as labels. The second model is a collaborative filtering AutoEncoder (AE) with the PCP and specialty procedures as input and output, respectively. The designed collaborative filtering AutoEncoder is similar to the the deep learning based collaborative models proposed in \cite{zhang2019deep, kuchaiev2017training}. The predictions from the base models are then fed into an ensemble neural network to improve the predictions from each of the base learners. Despite traditional ensembles methods that use the ratings from base learners to improve predictions \cite{moghimi2016boosted}, the proposed approach leverages the specialist id number as side information to personalize the recommendations both for the patient and speciality provider. Here, we develop and measure the potential advantages of the proposed method compared to clinical checklists and several other baselines.

\section{Cohort and Data Description}

In this work, we address the prediction of future clinical diagnostic steps for the outpatients referred to Stanford Health Care Endocrinology Clinic between Jan 2008 and Dec 2018. To have adequate access to the patients' clinical records, we only consider those referred by a PCP within Stanford Health Care Alliance network, which totally includes $6511$ patients. We aimed to predict the procedures (primarily lab and imaging tests) the endocrinologist would order at their first in-person visit. 
Because the procedures ordered could depend on the time window between the referral and the first specialist visit, we restricted the cohort to only those patients with a first specialist visit within 4 months after referral.

For each patient in our cohort we used electronic health record (EHR) data to extract all of the lab results within two months before the referral as well as the procedures ordered by the referring PCP. We further include the receiving specialist's identify (specialist ID) as  side information to allow the model to personalize predictions per specialist as well. 


\section{Proposed Method}
The proposed method is an ensemble model that takes the patient's clinical information and the specialist ID as input and predicts the future procedures. In order to feed the data into the model and train the base and ensemble models, we need to pre-process the data to the appropriate format. 

\subsection{Data Pre-Processing}\label{sec:data_process}
The defined cohort includes $6511$ patients and, within the defined cohort, there are $2993$ unique labs, $2158$ unique procedures, and $11810$ unique diagnosis codes. Given that it would not be practical to train a model with several thousand data dimensions and output labels using only $6511$ samples, we restricted each data category to only top most frequent types. Specifically, we only considered the top $100$ most common labs and top $60$ procedures. We also restricted the diagnosis codes to $10$ top most-prevalent codes related to endocrinology: \textit{Diabetes mellitus Type I or II, Hypercalcemia, Hyperlipidemia, Hypothyroidism, Hyperthyroidism, Osteopenia, Thyroid cancer, Thyroid nodule, and Obesity}. The raw lab results in the EHR data are mainly continuous data, which we converted into one-hot encoded format using the clinical laboratory defined "normal range" for each value. Thus, each lab value is embedded into a three dimensional binary vector, where the first dimension represents whether the lab value is available for the patient and the second and third dimensions indicate whether the lab value is low or high (in case of a normal result both are $0$). Thus, if a patient has any missing clinical information, the one hot encoding approach appropriately considers it the the encoded data format. Finally, the samples are randomly shuffled and split into the train and test sets with $80\%$ and $20\%$ of the entire sample sizes, respectively.

\subsection{Ensemble Model}\label{sec:method}

\begin{figure}
  \includegraphics[width=0.99\linewidth]{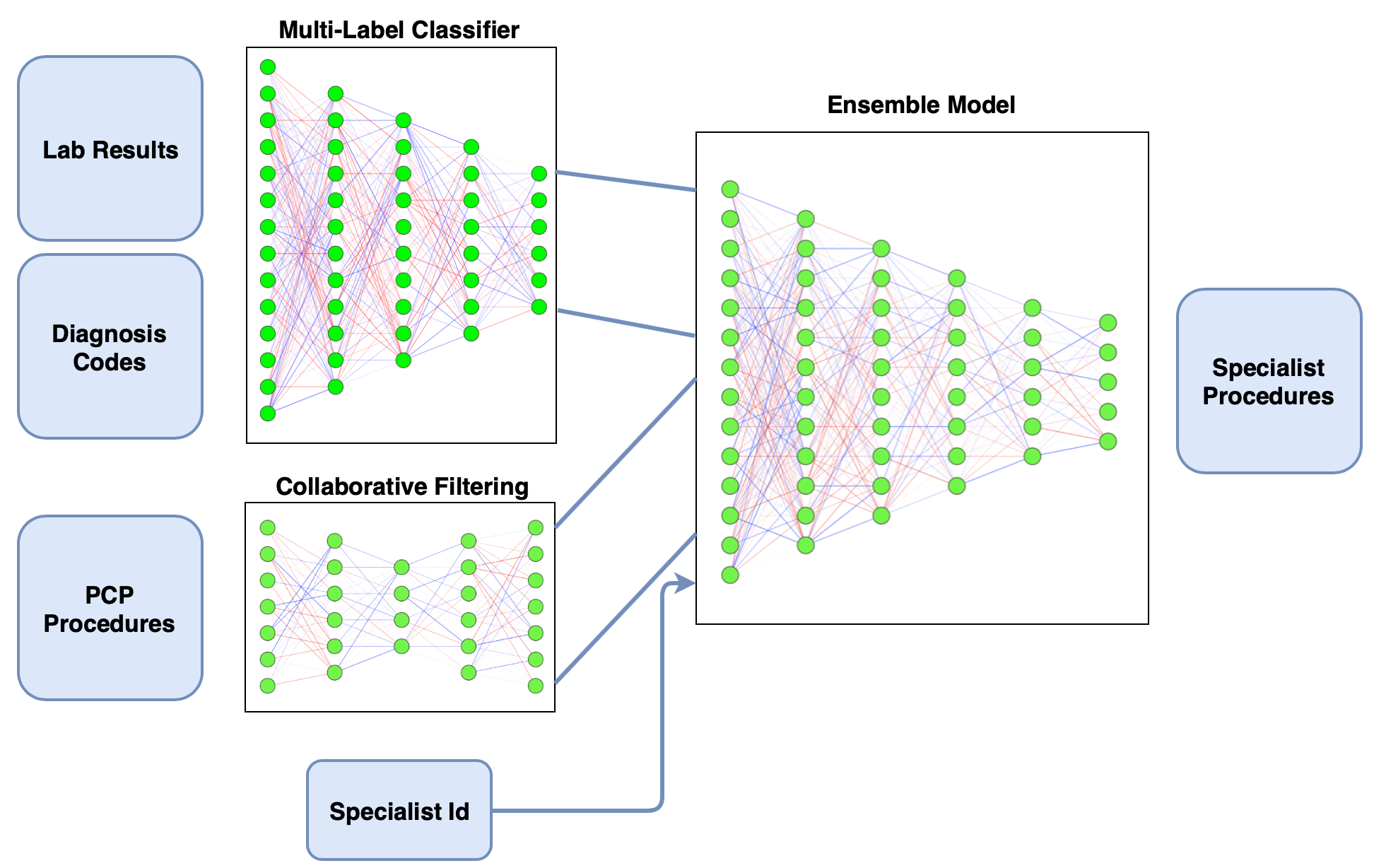}
  \caption{The proposed model consists of two base models which are trained separately and an ensemble model that combines the prediction results from the base models using the trained neural network.}
  \label{fig:model}
\end{figure}

The proposed model consists of two base models which are trained separately and an ensemble model that combines the prediction results from the base models using the trained neural network (Figure \ref{fig:model}). 
The first base model is a neural network based multi-label classifier with diagnostic data as input and specialty procedures as labels. The neural network consists of $5$ fully connected layers with the dimensions $310-200-100-80-60$ and rectified linear unit (ReLU) activations. The network is trained using stochatic gradient descent (SGD) with the learning rate $0.001$ and mean square error (MSE) loss function. The network is trained for $400$ epochs with the batch size of $256$. After each layer, a dropout regularization with $p=0.3$ is used to prevent overfitting. We refer to this network as diagnostic model (abbreviated as DM).
The second base model is an AutoEncoder (AE) based collaborative filtering architecture with the PCP and specialty procedures as input and output. The AE consists of $5$ fully connected layers with  dimensions $60-60-40-60-60$.
The predictions from the base models are then fed into an ensemble neural network which includes the specialist ID as side information to get the final predicted specialty procedures. The ensemble neural network consists of $6$ fully connected layers with the dimensions $130-200-150-100-80-60$ and each output neuron represents the score for a procedure ID. For all of the neural network based methods we performed several hyperparameter optimizations. The scores are normalized within the range $[0,1]$ and could be interpreted as an uncalibrated probability that the corresponding procedure is ordered by the specialist. 
Based on the predicted scores for the procedures we can take
two different recommendation approaches. The first method applies a fixed threshold and if the score of a given procedure is above the threshold, that procedure is recommended. Therefore, for different patients different numbers of procedures may be recommended. In the second approach the algorithm always recommends the top $k$ procedures. Thus, in this approach only the order of the scores are important not their values. In all of our experiments we used the recommendation based on a fixed score threshold since it resulted a better performance (reported in the Results section).


\section{Experiment Design}

The problem of predicting the specialty procedures using the lab results, diagnosis codes, and the PCP procedures is, in general, a multi-label classification problem and recommender system methods cannot be directly applied. However, we can split the clinical data into two major groups such that such that we can separately apply a multi-label classification model to the first group (lab results and diagnosis codes) and a collaborative filtering model to the second group (PCP procedures), which is of the same type as the output labels (Specialty procedures). We compared the results to two of the standard collaborative filtering method, i.e. singular value decomposition (SVD) and probabilistic matrix factorization (PMF).   
 We also compared the performance of the proposed ensemble method to each of the base models, i.e., the diagnostic model (DM) and AutoEncoder (AE), as well as the collaborative filtering methods SVD and PMF, and also a conventional clinical checklist. The clinical checklist was mainly retrieved and reviewed by our clinical author Ivana Jankovic from clinical guideline documents (UpToDate.com) for each of the main referral diagnoses to collate a checklist of relevant diagnostic procedure orders that should be considered for each.
 We also compared the results to an aggregate multi-label classifier based on neural networks (abbreviated as ANN in the figures) with $6$ fully connected layers which utilizes all the lab results, diagnosis codes, PCP procedures and specialist ID as a unified input and predicts the specialist-ordered procedures.


\section{Results}

\begin{figure}\centering
\includegraphics[width=0.99\linewidth]{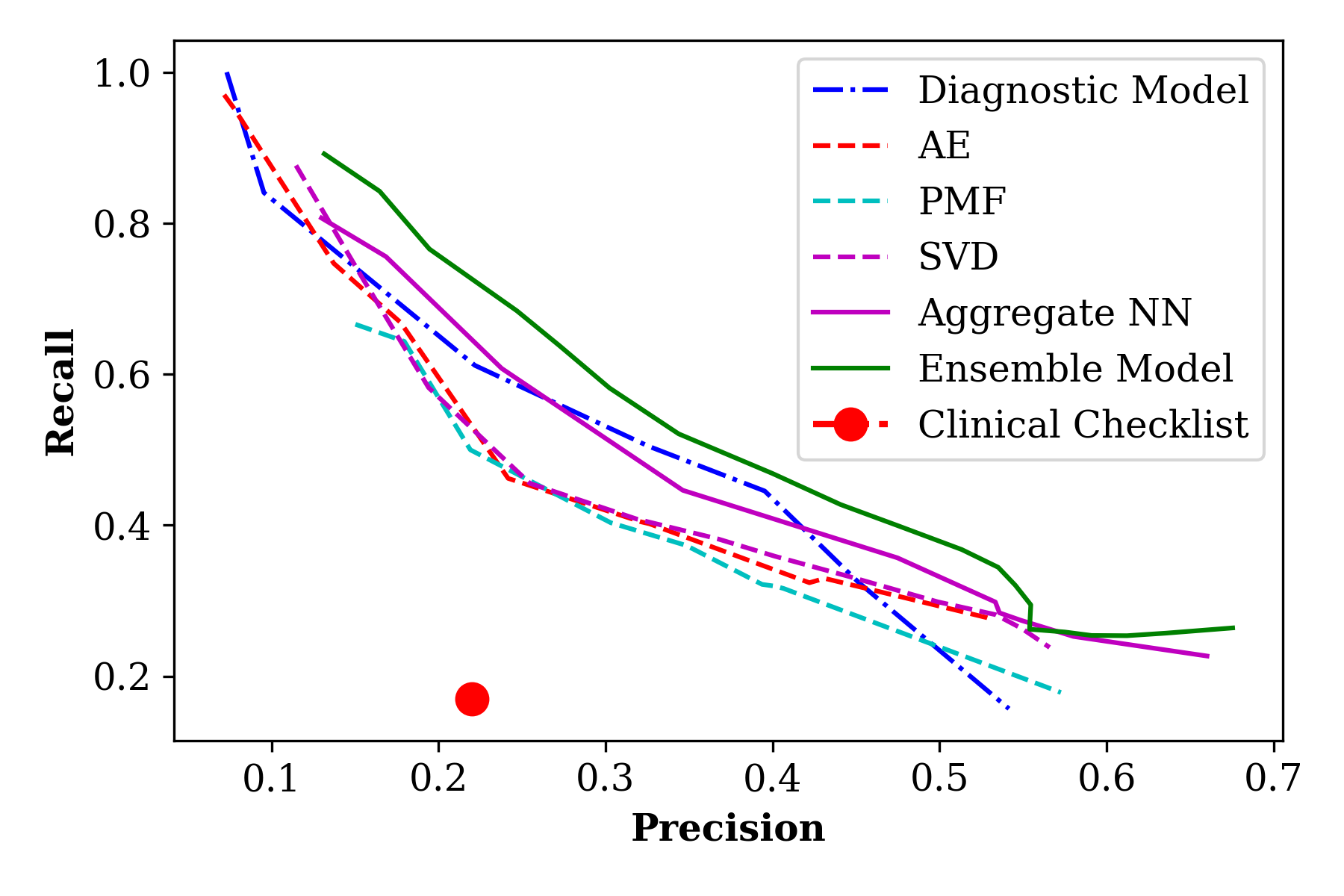}
\caption{Precision-Recall graph of the proposed ensemble method compared to the base models (diagnostic model and AutoEncoder), collaborative filtering methods (SVD and PMF), the aggregate neural network model (ANN), and clinical checklist.\label{fig:PR_graph}}
\end{figure}

\begin{figure}\centering
\includegraphics[width=0.99\linewidth]{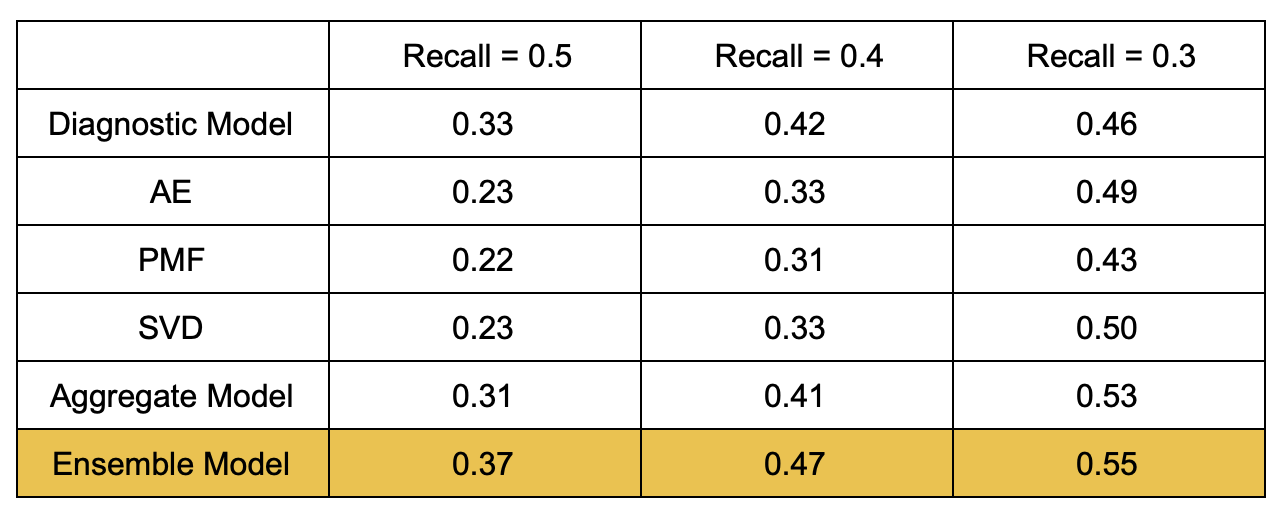}
\caption{Precision at fixed recall for the proposed ensemble method compared to the base models (diagnostic model and AE), collaborative filtering methods (SVD and PMF), the aggregate neural network model (ANN).  \label{fig:PR_table}}
\end{figure}

By varying the score threshold for each of the prediction methods to convert predicted scores into binary predictions for each procedure order, we can obtain different performance metrics including precision (positive predictive value, the fraction of predicted procedure orders the specialist actually ordered) and recall (sensitivity, the fraction of orders the specialist actually ordered that were predicted).
Therefore, the methods are evaluated in terms of precision, recall, and area under the receiver operating curve (AUROC) metrics.
Figure \ref{fig:PR_graph} represents the precision-recall graph of the proposed ensemble method compared to the base models (diagnostic model and AutoEncoder), collaborative filtering methods (SVD and PMF), the aggregate neural network model (ANN), and a clinical checklist. Precision at different fixed values of recall are represented in Figure \ref{fig:PR_table}. The ensemble method achieves a better precision-recall trade-off compared to the other models. The methods are also compared in terms of AUROC. The ensemble method achieves the highest AUROC of $0.80$ compared to the other methods.


\begin{figure}\centering
\includegraphics[width=0.99\linewidth]{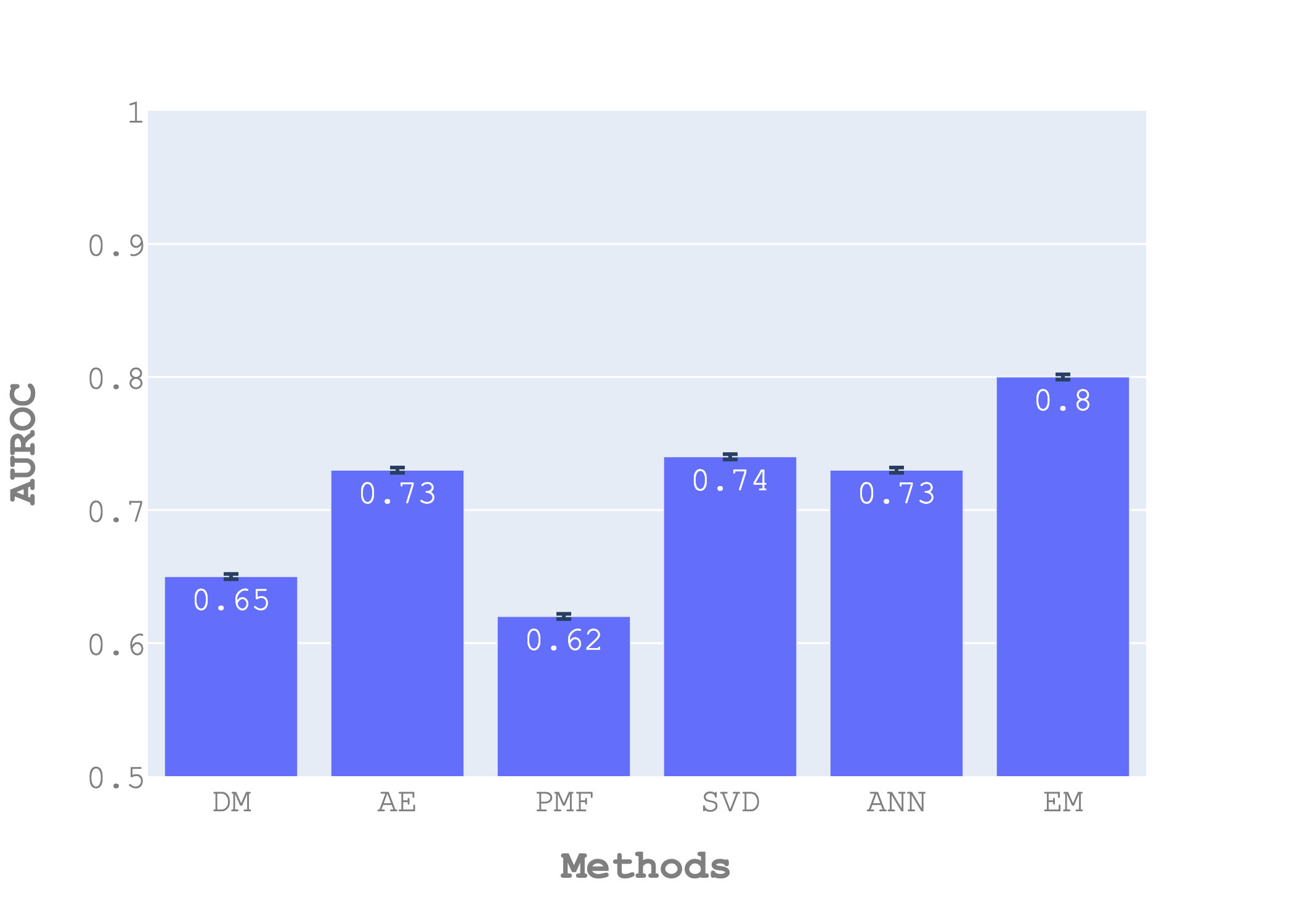}
\caption{AUROC of the proposed ensemble method (EM) compared to the base models (diagnostic model and AE), collaborative filtering methods (SVD and PMF), the aggregate neural network model (ANN). Error bars show the $95\%$ confidence interval computed using bootstrapped resampling. \label{fig:AUC}}
\end{figure}

\begin{figure}\centering
\includegraphics[width=0.99\linewidth]{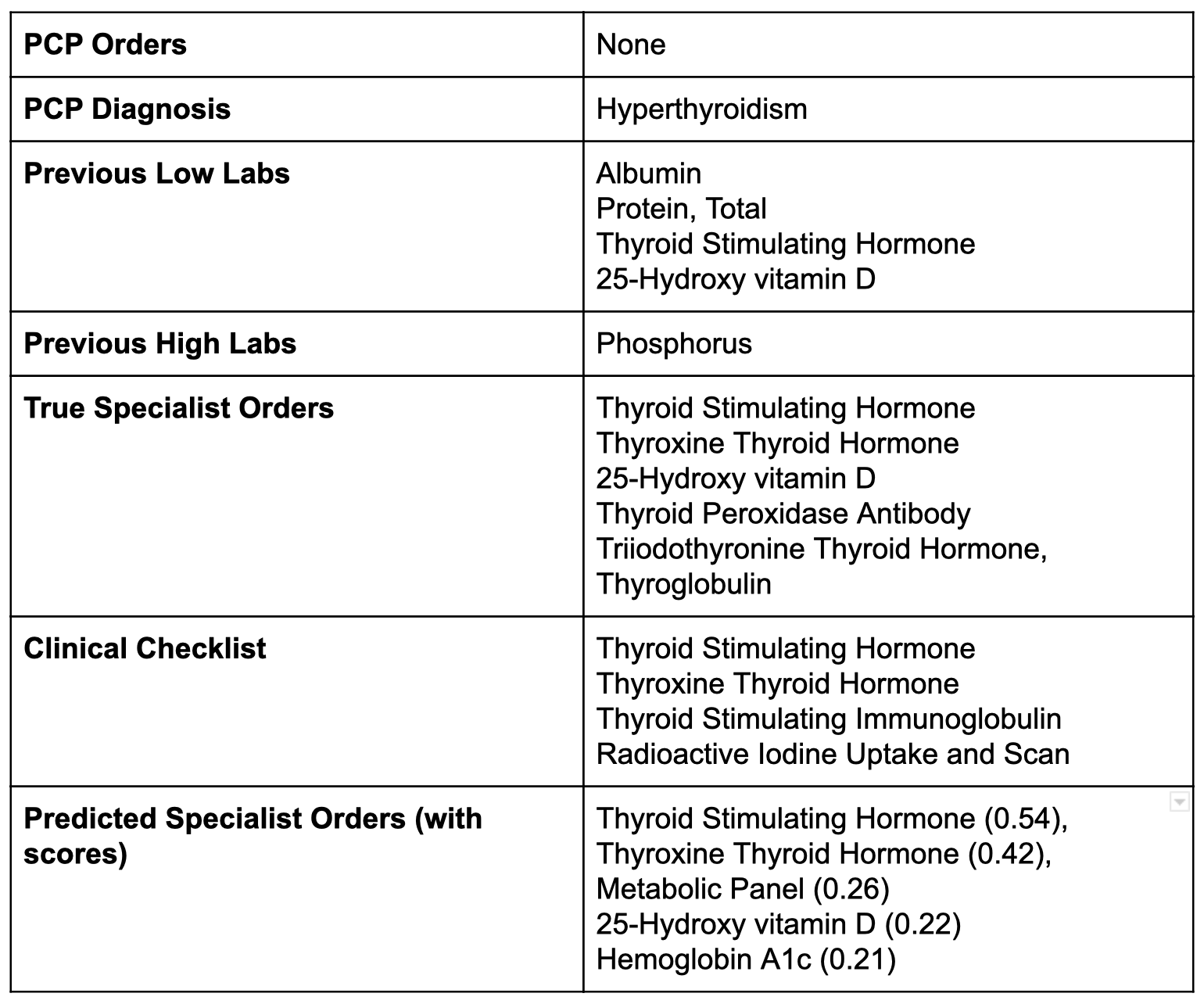}
\caption{A real-world example of a patient with the true specialist orders, the predicted procedures based on clinical checklist and the proposed ensemble method. \label{fig:example}}
\end{figure}
\begin{figure}\centering
\includegraphics[width=0.99\linewidth]{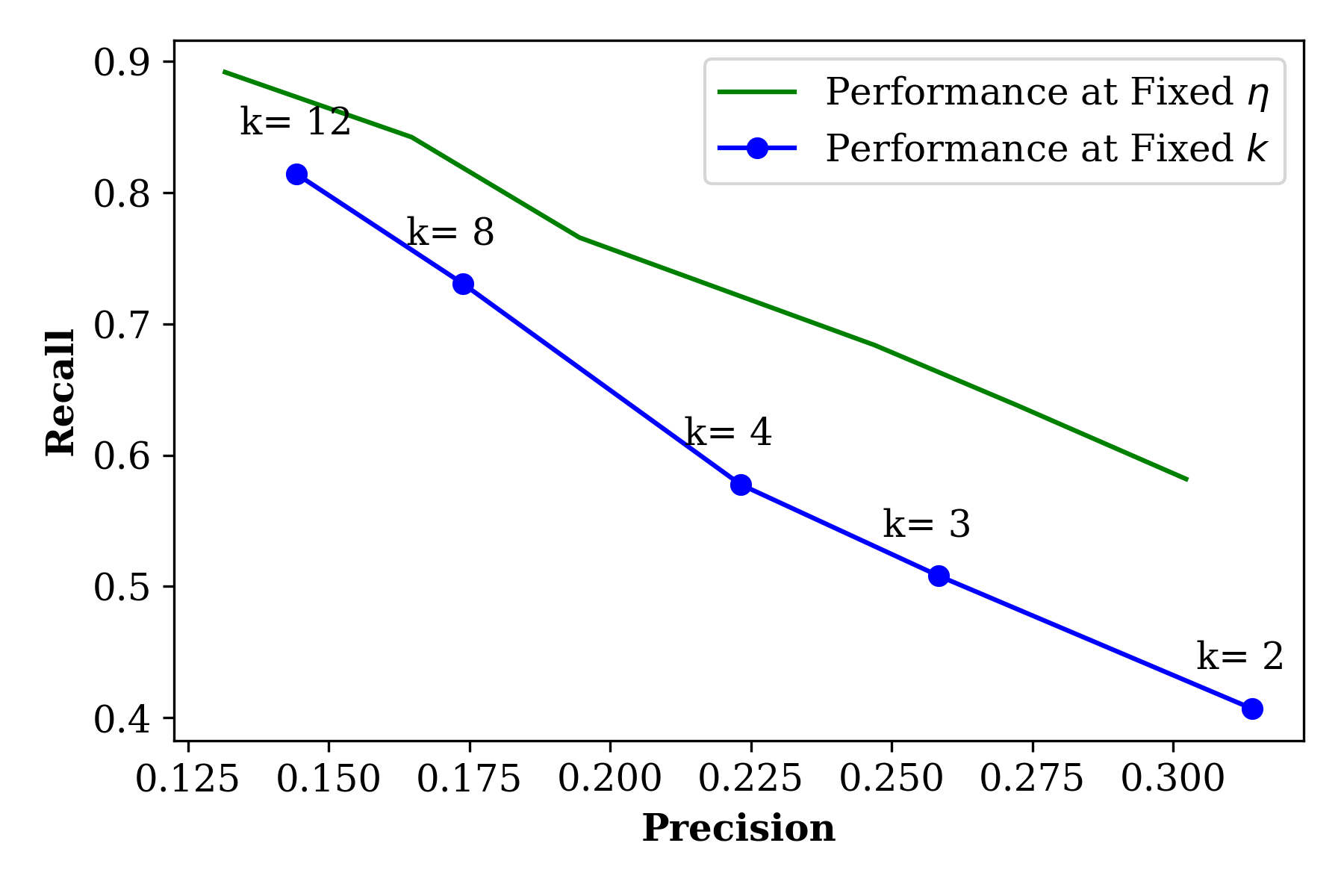}
\caption{Performance comparison of two different recommendation approaches based on the predicted scores for the ensemble method.  \label{fig:performance_at}}
\end{figure}

Figure \ref{fig:example} Example model inputs and outputs. Example patient's data up to time of speciality referral, the actual subsequent specialist procedure orders vs. predicted procedure orders from a diagnosis-based clinical checklist or predicted from our proposed ensemble method with a score threshold $0.20$. Finally we compare the performance of the ensemble method using two selection approaches based on the predicted scores (discussed in Section \ref{sec:method}). As shown in \ref{fig:performance_at} , the selection method based on a fixed threshold ($\eta$) performs better than the selection method based on the fixed $k$. 


\section{Discussion}
The generalizability of the proposed model to more diverse types of patients with different conditions depends on several key assumptions. As mentioned in \ref{sec:data_process}, due to the model's learning limitations with respect to the number of patients, we had to only include a portion of the labs, diagnosis codes and procedures as features and labels in our data, which degrades the performance of the recommendation model. Further, the recommended items based on the prediction model is learned based on the specialists' preferences, and they don't necessarily mean to be correct or incorrect orders. 




\section{Conclusion}

In this work we addressed the problem of predicting outpatient specialty diagnostic workups, specifically the procedure orders for diagnostic orders for adult Endocrinology referrals. We proposed a data-driven model that recommends follow-up procedure orders based on patients' clinical information.
Several evaluations illustrate that the proposed method can outperform conventional clinical checklist and baseline methods.

\bibliographystyle{ACM-Reference-Format}
\bibliography{ref}


\begin{thebibliography}{14}


\ifx \showCODEN    \undefined \def \showCODEN     #1{\unskip}     \fi
\ifx \showDOI      \undefined \def \showDOI       #1{#1}\fi
\ifx \showISBNx    \undefined \def \showISBNx     #1{\unskip}     \fi
\ifx \showISBNxiii \undefined \def \showISBNxiii  #1{\unskip}     \fi
\ifx \showISSN     \undefined \def \showISSN      #1{\unskip}     \fi
\ifx \showLCCN     \undefined \def \showLCCN      #1{\unskip}     \fi
\ifx \shownote     \undefined \def \shownote      #1{#1}          \fi
\ifx \showarticletitle \undefined \def \showarticletitle #1{#1}   \fi
\ifx \showURL      \undefined \def \showURL       {\relax}        \fi
\providecommand\bibfield[2]{#2}
\providecommand\bibinfo[2]{#2}
\providecommand\natexlab[1]{#1}
\providecommand\showeprint[2][]{arXiv:#2}

\bibitem[\protect\citeauthoryear{Chen, Goldstein, Asch, Mackey, and
  Altman}{Chen et~al\mbox{.}}{2017}]%
        {chen2017predicting}
\bibfield{author}{\bibinfo{person}{Jonathan~H Chen}, \bibinfo{person}{Mary~K
  Goldstein}, \bibinfo{person}{Steven~M Asch}, \bibinfo{person}{Lester Mackey},
  {and} \bibinfo{person}{Russ~B Altman}.} \bibinfo{year}{2017}\natexlab{}.
\newblock \showarticletitle{Predicting inpatient clinical order patterns with
  probabilistic topic models vs conventional order sets}.
\newblock \bibinfo{journal}{\emph{Journal of the American Medical Informatics
  Association}} \bibinfo{volume}{24}, \bibinfo{number}{3}
  (\bibinfo{year}{2017}), \bibinfo{pages}{472--480}.
\newblock


\bibitem[\protect\citeauthoryear{Choi, Bahadori, Schuetz, Stewart, and
  Sun}{Choi et~al\mbox{.}}{2016}]%
        {choi2016doctor}
\bibfield{author}{\bibinfo{person}{Edward Choi}, \bibinfo{person}{Mohammad~Taha
  Bahadori}, \bibinfo{person}{Andy Schuetz}, \bibinfo{person}{Walter~F
  Stewart}, {and} \bibinfo{person}{Jimeng Sun}.}
  \bibinfo{year}{2016}\natexlab{}.
\newblock \showarticletitle{Doctor ai: Predicting clinical events via recurrent
  neural networks}. In \bibinfo{booktitle}{\emph{Machine Learning for
  Healthcare Conference}}. \bibinfo{pages}{301--318}.
\newblock


\bibitem[\protect\citeauthoryear{Giger}{Giger}{2018}]%
        {giger2018machine}
\bibfield{author}{\bibinfo{person}{Maryellen~L Giger}.}
  \bibinfo{year}{2018}\natexlab{}.
\newblock \showarticletitle{Machine learning in medical imaging}.
\newblock \bibinfo{journal}{\emph{Journal of the American College of
  Radiology}} \bibinfo{volume}{15}, \bibinfo{number}{3} (\bibinfo{year}{2018}),
  \bibinfo{pages}{512--520}.
\newblock


\bibitem[\protect\citeauthoryear{Hao and Blair}{Hao and Blair}{2016}]%
        {hao2016comparative}
\bibfield{author}{\bibinfo{person}{Fang Hao} {and}
  \bibinfo{person}{Rachael~Hageman Blair}.} \bibinfo{year}{2016}\natexlab{}.
\newblock \showarticletitle{A comparative study: classification vs. user-based
  collaborative filtering for clinical prediction}.
\newblock \bibinfo{journal}{\emph{BMC medical research methodology}}
  \bibinfo{volume}{16}, \bibinfo{number}{1} (\bibinfo{year}{2016}),
  \bibinfo{pages}{172}.
\newblock


\bibitem[\protect\citeauthoryear{Kenny, Parsons, Gratch, and Rizzo}{Kenny
  et~al\mbox{.}}{2008}]%
        {kenny2008virtual}
\bibfield{author}{\bibinfo{person}{Patrick Kenny}, \bibinfo{person}{Thomas
  Parsons}, \bibinfo{person}{Jonathan Gratch}, {and} \bibinfo{person}{Albert
  Rizzo}.} \bibinfo{year}{2008}\natexlab{}.
\newblock \showarticletitle{Virtual humans for assisted health care}. In
  \bibinfo{booktitle}{\emph{Proceedings of the 1st international conference on
  PErvasive Technologies Related to Assistive Environments}}.
  \bibinfo{pages}{1--4}.
\newblock


\bibitem[\protect\citeauthoryear{Komkhao and Halang}{Komkhao and
  Halang}{2013}]%
        {komkhao2013recommender}
\bibfield{author}{\bibinfo{person}{Maytiyanin Komkhao} {and}
  \bibinfo{person}{Wolfgang~A Halang}.} \bibinfo{year}{2013}\natexlab{}.
\newblock \showarticletitle{Recommender systems in telemedicine}.
\newblock \bibinfo{journal}{\emph{IFAC Proceedings Volumes}}
  \bibinfo{volume}{46}, \bibinfo{number}{28} (\bibinfo{year}{2013}),
  \bibinfo{pages}{28--33}.
\newblock


\bibitem[\protect\citeauthoryear{Kuchaiev and Ginsburg}{Kuchaiev and
  Ginsburg}{2017}]%
        {kuchaiev2017training}
\bibfield{author}{\bibinfo{person}{Oleksii Kuchaiev} {and}
  \bibinfo{person}{Boris Ginsburg}.} \bibinfo{year}{2017}\natexlab{}.
\newblock \showarticletitle{Training deep autoencoders for collaborative
  filtering}.
\newblock \bibinfo{journal}{\emph{arXiv preprint arXiv:1708.01715}}
  (\bibinfo{year}{2017}).
\newblock


\bibitem[\protect\citeauthoryear{Lakshmanaprabu, Mohanty, Krishnamoorthy,
  Uthayakumar, Shankar, et~al\mbox{.}}{Lakshmanaprabu et~al\mbox{.}}{2019}]%
        {lakshmanaprabu2019online}
\bibfield{author}{\bibinfo{person}{SK Lakshmanaprabu},
  \bibinfo{person}{Sachi~Nandan Mohanty}, \bibinfo{person}{Sujatha
  Krishnamoorthy}, \bibinfo{person}{J Uthayakumar}, \bibinfo{person}{K
  Shankar}, {et~al\mbox{.}}} \bibinfo{year}{2019}\natexlab{}.
\newblock \showarticletitle{Online clinical decision support system using
  optimal deep neural networks}.
\newblock \bibinfo{journal}{\emph{Applied Soft Computing}}
  \bibinfo{volume}{81} (\bibinfo{year}{2019}), \bibinfo{pages}{105487}.
\newblock


\bibitem[\protect\citeauthoryear{Merdan, Ghani, and Denton}{Merdan
  et~al\mbox{.}}{2018}]%
        {merdan2018integrating}
\bibfield{author}{\bibinfo{person}{Selin Merdan}, \bibinfo{person}{Khurshid
  Ghani}, {and} \bibinfo{person}{Brian Denton}.}
  \bibinfo{year}{2018}\natexlab{}.
\newblock \showarticletitle{Integrating Machine Learning and Optimization
  Methods for Imaging of Patients with Prostate Cancer}. In
  \bibinfo{booktitle}{\emph{Machine Learning for Healthcare Conference}}.
  \bibinfo{pages}{119--136}.
\newblock


\bibitem[\protect\citeauthoryear{Middleton, Sittig, and Wright}{Middleton
  et~al\mbox{.}}{2016}]%
        {middleton2016clinical}
\bibfield{author}{\bibinfo{person}{B Middleton}, \bibinfo{person}{DF Sittig},
  {and} \bibinfo{person}{A Wright}.} \bibinfo{year}{2016}\natexlab{}.
\newblock \showarticletitle{Clinical decision support: a 25 year retrospective
  and a 25 year vision}.
\newblock \bibinfo{journal}{\emph{Yearbook of medical informatics}}
  \bibinfo{volume}{25}, \bibinfo{number}{S 01} (\bibinfo{year}{2016}),
  \bibinfo{pages}{S103--S116}.
\newblock


\bibitem[\protect\citeauthoryear{Miotto, Li, Kidd, and Dudley}{Miotto
  et~al\mbox{.}}{2016}]%
        {miotto2016deep}
\bibfield{author}{\bibinfo{person}{Riccardo Miotto}, \bibinfo{person}{Li Li},
  \bibinfo{person}{Brian~A Kidd}, {and} \bibinfo{person}{Joel~T Dudley}.}
  \bibinfo{year}{2016}\natexlab{}.
\newblock \showarticletitle{Deep patient: an unsupervised representation to
  predict the future of patients from the electronic health records}.
\newblock \bibinfo{journal}{\emph{Scientific reports}} \bibinfo{volume}{6},
  \bibinfo{number}{1} (\bibinfo{year}{2016}), \bibinfo{pages}{1--10}.
\newblock


\bibitem[\protect\citeauthoryear{Moghimi, Belongie, Saberian, Yang,
  Vasconcelos, and Li}{Moghimi et~al\mbox{.}}{2016}]%
        {moghimi2016boosted}
\bibfield{author}{\bibinfo{person}{Mohammad Moghimi}, \bibinfo{person}{Serge~J
  Belongie}, \bibinfo{person}{Mohammad~J Saberian}, \bibinfo{person}{Jian
  Yang}, \bibinfo{person}{Nuno Vasconcelos}, {and} \bibinfo{person}{Li-Jia
  Li}.} \bibinfo{year}{2016}\natexlab{}.
\newblock \showarticletitle{Boosted Convolutional Neural Networks.}. In
  \bibinfo{booktitle}{\emph{BMVC}}, Vol.~\bibinfo{volume}{5}.
  \bibinfo{pages}{6}.
\newblock


\bibitem[\protect\citeauthoryear{Prentice and Pizer}{Prentice and
  Pizer}{2007}]%
        {prentice2007delayed}
\bibfield{author}{\bibinfo{person}{Julia~C Prentice} {and}
  \bibinfo{person}{Steven~D Pizer}.} \bibinfo{year}{2007}\natexlab{}.
\newblock \showarticletitle{Delayed access to health care and mortality}.
\newblock \bibinfo{journal}{\emph{Health services research}}
  \bibinfo{volume}{42}, \bibinfo{number}{2} (\bibinfo{year}{2007}),
  \bibinfo{pages}{644--662}.
\newblock


\bibitem[\protect\citeauthoryear{Zhang, Yao, Sun, and Tay}{Zhang
  et~al\mbox{.}}{2019}]%
        {zhang2019deep}
\bibfield{author}{\bibinfo{person}{Shuai Zhang}, \bibinfo{person}{Lina Yao},
  \bibinfo{person}{Aixin Sun}, {and} \bibinfo{person}{Yi Tay}.}
  \bibinfo{year}{2019}\natexlab{}.
\newblock \showarticletitle{Deep learning based recommender system: A survey
  and new perspectives}.
\newblock \bibinfo{journal}{\emph{ACM Computing Surveys (CSUR)}}
  \bibinfo{volume}{52}, \bibinfo{number}{1} (\bibinfo{year}{2019}),
  \bibinfo{pages}{1--38}.
\newblock


\end{thebibliography}

\end{document}